\documentclass[11pt]{article}
\usepackage[margin=1.2in]{geometry}
\usepackage{amsmath, amsthm, amssymb}
\usepackage{booktabs}
\usepackage{graphicx}
\usepackage{xcolor}
\usepackage{hyperref}
\usepackage{xurl}
\usepackage{natbib}
\usepackage{multirow}
\usepackage{microtype}
\usepackage{booktabs}
\usepackage{parskip}
\usepackage{algorithm}
\usepackage{algorithmic}
\usepackage{doi}
\hypersetup{
  colorlinks=true,
  linkcolor=blue!70!black,
  citecolor=blue!70!black,
  urlcolor=blue!70!black
}

\newtheorem{prediction}{Prediction}
\newcommand{\eff}{\mathrm{eff}}
\newcommand{\erank}{\rho_{\eff}}
\newcommand{\drift}{\Delta_{\mathrm{CKA}}}

\newcommand{\R}{\mathbb{R}}

\newcommand{\Attn}{\mathcal{A}}
\title{\textbf{Relevant and Irrelevant:\\ A Renormalization Group Analysis of Transformer Attention}}
\author{
    Parviz Haggi-Mani\textsuperscript{1,2}, Irina Rish\textsuperscript{1,2}\\
    haggimpa@mila.quebec, irina.rish@mila.quebec\\
    \textsuperscript{1}\textnormal{Université de Montréal},
    \textsuperscript{2}\textnormal{Mila -- Quebec AI Institute}
}
\date{}
\begin{document}
\maketitle
\begin{abstract}
Using the language of Wilsonian renormalization group theory (RG),
we treat the Transformer's attention mechanism as a perturbation of
the trained MLP residual-stack fixed point and ask whether it
constitutes a relevant, marginal, or irrelevant operator.
We derive a fixed-point shift formula $\delta = -{M^*}^{-1}(a+b)$
and obtain four testable predictions for the fixed-point geometry,
effective rank profile, layer specificity, and perturbation decay
spectrum.
Testing these on synthetic Markov chain sequences with controlled
correlation length $\xi$, we find:
(1)~For long-$\xi$ chains, attention is strongly \emph{relevant}:
it closes a residual loss gap the MLP cannot bridge and drives a
phase transition in representation space, with effective rank
jumping above input dimensionality at layer~1 and stabilising at
a high-dimensional plateau.
(2)~For short-$\xi$ chains, attention is \emph{irrelevant}: the
Transformer converges to the same loss and fixed-point geometry as
the MLP, though Experiment~4 shows it contracts perturbations
faster.
(3)~The transition is dominated by the first-layer head (L0H0),
which accounts for more than $4\times$ the representational shift
of any subsequent head, consistent with the prediction that the
relevant operator acts before the MLP begins integrating out
positional variation.
(4)~Perturbation decay experiments reveal a regime reversal: in
the long-$\xi$ regime the Transformer selectively preserves slow
Markov modes ($5.4\times$ dynamic range in decay length vs.\
$1.3\times$ for the MLP); in the short-$\xi$ regime it suppresses
all modes faster than the MLP, with no spectral selectivity.
Together, these results show that the relevance of attention is
not a property of the architecture but of the spectral structure
of the data-generating process, and that a first-order RG
perturbation framework provides a predictive account of that
difference.
\end{abstract}
\section{Introduction}
\label{sec:intro}
 In our previous work \citep{haggimani2026phase1} we showed that a pure
MLP residual stack trained on masked token prediction over Markov chain
sequences implements a selective coarse-graining procedure governed by
the correlation length $\xi$ of the input distribution:
short-$\xi$ chains produce monotone rank collapse ($8.4\times$
compression), long-$\xi$ chains preserve the effective rank of the hidden representations,
and in both cases inter-layer kernel drift concentrates at one or two
transitions, with the remainder of the network near a
\emph{fixed-point plateau} consistent with RG theory.
One important finding in this controlled setting is that the forward
pass does not merely \emph{resemble} an RG flow; the sequence of
representations $(h^{(0)}, \ldots, h^{(L)})$ executes one, with
each layer performing one coarse-graining step and the fixed-point
plateau marking convergence to the attractor.
The entry transition is the point at which the network commits to a
particular attractor in representation space: the sharp concentration
of kernel drift at one or two layers reflects the discrete nature of
this flow, where the network crosses from one representational regime
to another in a single coarse-graining step rather than gradually.
The natural next step is to introduce attention and characterize its
effect on the RG flow. In this framework, attention enters as an
additional operator whose relevance or irrelevance determines whether
it shifts the fixed point by a small amount or drives the system to
a qualitatively different attractor.
In the language of statistical field theory, this is equivalent to
asking whether attention is a \emph{relevant} perturbation of the
MLP fixed point (grows under RG iteration with depth, drives the
system to a different fixed point), \emph{marginal} (preserved
under iteration, persists at constant amplitude), or
\emph{irrelevant} (decays, leaving the fixed-point physics
unchanged). The type of operator determines the large-scale representational
behavior of the network.
This builds on a line of work connecting RG to learning systems.
\citet{mehta2014exact} construct an exact mapping between Kadanoff's
variational RG and RBM-based deep networks, establishing the conceptual
foundation; \citet{coppola2026renormalization} extend this to weakly
non-linear networks, developing a rigorous RG framework that classifies
perturbations as relevant or irrelevant and reveals universality in
learning curves at large data limits. Neither work, however, derives or
tests perturbation-theoretic predictions for the fixed-point structure
of a trained network's representations \citep{bordelon2024renormalization}.
In the spirit of Kadanoff--Wilson RG \citep{wilson1971renormalization},
recent work has connected RG-like dynamics to topological phase
transitions in representation manifolds
\citep{alpay2026latentobjectpermanencetopological}.
\citet{makkuva2025attention} study Transformers trained on first-order
Markov chains and analytically characterize fixed points of the loss
landscape, showing that attention can drive the system between a
unigram and a bigram attractor, which is the closest methodological
antecedent to the present work in its combination of Markov input
statistics and fixed-point reasoning, though it does not use an RG
framework or classify attention as an operator.
\citet{fernando2026dynamics} provide empirical evidence that training
installs a monotonic spectral gradient through depth and that
perturbations are differentially amplified or suppressed layer by layer,
consistent with RG-like flow, but without formal fixed-point predictions.
The present paper derives the fixed-point shift formula
$\delta = -{M^*}^{-1}(a+b)$ and uses it to make falsifiable
predictions about the relevance of attention as a function of the
correlation length $\xi$ of the input distribution, testing those
predictions against trained network measurements.
Applying this framework requires a formal definition of what it
means for attention to perturb the MLP fixed point, and testable
predictions for how measured quantities should respond.
We provide both in Section~\ref{sec:theory}, then test the
predictions in Sections~\ref{sec:e1}--\ref{sec:e4}.
\paragraph{Paper structure.}
Section~\ref{sec:background} is a review of our previous setup in \citep{haggimani2026phase1}.
Section~\ref{sec:theory} develops the perturbation-theoretic framework
and derives testable predictions.
Sections~\ref{sec:e1}--\ref{sec:e4} present the experiments.
Section~\ref{sec:discussion} assesses the predictions against the
results, and Section~\ref{sec:conclusion} concludes. Some calculations are provided in the Appendix.
\section{Background}
\label{sec:background}
\subsection{Synthetic corpus}
Sequences are sampled from a Markov chain over vocabulary $V = 16$ with
row-stochastic transition matrix $P$.
The spectral gap governs the correlation length
$\xi = -1/\log|\lambda_2|$, where $\lambda_2$ is the second-largest
eigenvalue by magnitude.
Two regimes are studied:
\begin{itemize}
  \item \textbf{Short-$\xi$}: $\alpha = 10$,
        $\lambda_2 \approx 0.44$, $\xi \approx 1.2$.
        Sequences decorrelate within $\sim 5$ steps.
  \item \textbf{Long-$\xi$}: $\alpha = 0.05$,
        $\lambda_2 \approx 0.86$, $\xi \approx 6.7$.
        Mixing time $t_\text{mix}(0.01) \approx 31$ steps; the context
        window carries substantial predictive signal.
\end{itemize}
In all experiments $T = 64$ and sequences are encoded as one-hot vectors
$X \in \R^{B \times T \times V}$.
Training uses BERT-style masked token prediction (mask rate 15\%),
cross-entropy loss, Adam \citep{kingma2017adammethodstochasticoptimization}
with learning rate $3\times10^{-4}$, batch size $B=32$, for
10{,}000 steps with fresh sequences at each step.
\subsection{Architectures}
\paragraph{MLP baseline:}
A pre-norm MLP residual stack with no attention: input projection
$\phi_\text{in}: \R^V \to \R^d$, $L$ residual blocks each computing
$x \leftarrow x + \text{MLP}(\text{LayerNorm}(x))$ with hidden dimension
$4d$ and GELU activation, final LayerNorm, and classification head
$\phi_\text{head}: \R^d \to \R^V$.
All experiments use $d = 64$, $L = 6$.
\paragraph{TFM:}
A matched architecture in which each residual block adds a pre-norm
multi-head self-attention sublayer before the MLP sublayer.
No positional encoding is used, making attention the only new inductive
bias (see Section~\ref{sec:discussion}).
All other hyperparameters are identical to the MLP; $n_\text{heads} = 1$
in the main experiments.
\subsection{Measurement quantities}
\emph{Effective rank} \citep{roy2007effective}:
for a representation matrix $H \in \R^{N \times d}$ with normalized
singular-value spectrum $p_i = \sigma_i / \sum_j \sigma_j$,
\begin{equation}
  \erank(H) = \exp\!\left(-\sum_i p_i \log p_i\right) \in [1, d].
  \label{eq:erank}
\end{equation}
A decreasing depth profile signals progressive coarse-graining
\citep{haggimani2026phase1}, a pattern also observed empirically in
large pre-trained models \citep{alpay2026latentobjectpermanencetopological,
fernando2026dynamics}.
\emph{Kernel drift:}
$\drift(l, l{+}1) = 1 - \text{CKA}(H^{(l)}, H^{(l+1)})$,
where CKA is centered kernel alignment \citep{kornblith2019similarity}
\begin{equation}
    \mathrm{CKA}(X, Y) = \frac{\|Y^\top X\|_F^2}{\|X^\top X\|_F \|Y^\top Y\|_F},
\end{equation}
and $X, Y \in \mathbb{R}^{n \times d}$ are centered representation
matrices over $n$ examples:
Small drift indicates a fixed point; concentrated drift marks discrete
transition events.
The layer-by-layer progression from syntactic to semantic representations
observed in BERT \citep{tenney2019bert} is consistent with the
coarse-graining interpretation adopted here.
All representations are extracted in flatten mode ($N = B \times T$)
to avoid the mean-pooling artifact identified in
\citep{haggimani2026phase1}.
\section{Theoretical Framework: Attention as an RG Perturbation}
\label{sec:theory}
\subsection{The stability of the MLP fixed point}
\label{sec:theory_fixedpt}
As established in our previous work, the trained MLP residual stack converges to a
fixed-point plateau: after one or two entry transitions, inter-layer
kernel drift is near zero and representations undergo no further geometric
change. The stability of this fixed-point (See Appendix~\ref{appendix:MLP_fp}) is governed by
\begin{equation}
\label{eq:mlp_stability}
    \epsilon_n = (I + {M})^n\,\epsilon_0,
\end{equation}
where $n$ is the number of iterations. In the eigenbasis of the Jacobian ${M}$ with eigenvalues $\mu_k$ this is
\begin{equation}
    [\epsilon_n]_k = (1 + \mu_k)^n\,[\epsilon_0]_k.
\end{equation}
The MLP fixed-point is stable if and only if all eigenvalues of
$I + {M}$ lie strictly inside the unit circle:
\begin{equation}
    |1 + \mu_k| < 1 \quad \text{for all } k.
\end{equation}
For real eigenvalues this reduces to $-2 < \mu_k < 0$: the MLP
sub-network must contract perturbations at $x^*$, but not so strongly
as to overshoot. Modes with $\mu_k > 0$ or $\mu_k < -2$ grow under iteration (unstable eigenvector directions).
The fixed-point plateau observed empirically in both $\xi$ regimes
implies that the trained MLP satisfies this condition across all modes.
The low effective rank ($\erank \approx 1.8$) of the attractor reflects
that most eigenvalues $\mu_k$ are close to $-1$, so that nearly all
directions in representation space are contracted to zero, leaving only
a low-dimensional stable subspace.
\subsection{The TFM Fixed-Point as a Perturbation of the MLP Fixed-Point}
\label{sec:theory_pert}
We add a single-head attention sublayer
\begin{equation}
 \Attn(x) = \mathrm{softmax}(QK^\top/\sqrt{d})\,V
\end{equation}
with
$Q=xW_Q$, $K=xW_K$, $V=xW_V$ to each MLP block.
Assuming that the new fixed point $\tilde{x}^*$ is a small shift $\delta$ away from the previous MLP fixed point (See Appendix \ref{appendix:fp_shift}), the shift is given by
\begin{equation}
  \delta \;=\; -{M^*}^{-1}(a + b)
  \label{eq:fp_shift}
\end{equation}
where $M^*\equiv (I+D[b])(I+D[a])-I$, i.e.\
\begin{equation}
    M^* = D[b] + D[a] + D[b]D[a],
    \label{eq:M_star}
\end{equation}
and we have defined
\begin{equation}
  a \equiv \Attn(x^*), \quad
  b \equiv f(x^* + a), \quad
  D[a]\equiv D[\Attn](x^*) , \quad
  D[b]\equiv D[f](x^*+a).
  \end{equation}
 
We have assumed in formula~\eqref{eq:fp_shift} that ${M^*}$ is invertible.
Note that $a + b = 0$ if and only if $x^*$ is already a fixed point of
the modified block; hence $a + b$ measures the residual of the
fixed-point condition evaluated at $x^*$.
 Decomposing $a + b$ in the eigenbasis $\{v_k\}$ of ${M^*}$
with eigenvalues $\{\nu_k\}$, formula~\eqref{eq:fp_shift} gives
\begin{equation}
\label{eq:shift_nu}
    \delta = -\sum_k \frac{[a+b]_k}{\nu_k}\, v_k,\quad [\delta]_k = -[a+b]_k/\nu_k,
\end{equation}
where $[\delta]_k$ is the shift along the eigenvector $v_k$.
\textbf{The Stability of the TFM Fixed-Point}
(see Appendix~\ref{appendix:tfm_stability}).
Linearizing around $\tilde{x}^*$ with $x_n = \tilde{x}^* + \epsilon_n$,
the perturbation evolves as
\begin{equation}
\label{eq:tfm_stability}
    \epsilon_n = \tilde{M}^{*n}\,\epsilon_0,
    \qquad
    \tilde{M}^* \equiv (I+D[\tilde{b}])(I+D[\tilde{a}]),
\end{equation}
where
\begin{equation}
    \tilde{a} \equiv \Attn(\tilde{x}^*), \quad
    \tilde{b} \equiv f(\tilde{x}^*+\tilde{a}) = -\tilde{a}, \quad
    D[\tilde{a}] \equiv D[\Attn](\tilde{x}^*), \quad
    D[\tilde{b}] \equiv D[f](\tilde{x}^*+\tilde{a}).
\end{equation}
The fixed point $\tilde{x}^*$ is stable if and only if all eigenvalues
$\tilde{\nu}_k$ of $\tilde{M}^*$ lie strictly inside the unit circle:
\begin{equation}
    |\tilde{\nu}_k| < 1 \quad \text{for all } k.
\end{equation}
In the eigenbasis $\{\tilde{v}_k\}$ of $\tilde{M}^*$ each mode evolves as
$[\epsilon_n]_k = \tilde{\nu}_k^n\,[\epsilon_0]_k$, so modes with
$|\tilde{\nu}_k| < 1$ decay, modes with $|\tilde{\nu}_k| > 1$ grow,
and modes with $|\tilde{\nu}_k| = 1$ are marginal.
\textbf{Note} that while $\tilde{M}^*$ ($\{\tilde v_k,\tilde \nu_k\}$) is evaluated at the TFM fixed point $\tilde{x}^*$, the $M^*$ ($\{v_k,\nu_k\}$) of the shift formula~\eqref{eq:M_star} is evaluated at
the MLP fixed point $x^*$.
However, when the shift $\delta = \tilde{x}^* - x^*$ is small,  i.e. the two
evaluation points are close
\begin{equation}
    D[\tilde{a}] \approx D[a], \qquad D[\tilde{b}] \approx D[b],
\end{equation}
so that
\begin{equation}
    \tilde{M}^* = (I+D[\tilde{b}])(I+D[\tilde{a}])
    \approx (I+D[b])(I+D[a]) = I + M^*.
\end{equation}
In this regime the stability condition $|\tilde{\nu}_k|<1$ reduces to
$|1+\nu_k|<1$ on the eigenvalues $\nu_k$ of $M^*$, i.e.\
$-2 < \nu_k < 0$ for real eigenvalues, the same condition as for
the MLP fixed point, but with the Jacobian now evaluated at the
attention-shifted point $x^* + a$.
This connects the shift formula and the stability analysis: under
the approximation $\tilde{M}^* \approx I + M^*$, the same matrix
$M^*$ that controls the magnitude of the fixed-point shift
$\delta = -{M^*}^{-1}(a+b)$ also governs the stability of the new
fixed point.
In particular, near a bifurcation where $M^*$ develops a
near-zero eigenvalue $\nu_k \approx 0$, the shift $\delta$ diverges
along $v_k$ while simultaneously $\tilde{\nu}_k = 1 + \nu_k \to 1$,
pushing that mode to the boundary of the unit circle and rendering the
fixed point marginally stable.
\textbf{Note} that equations~\eqref{eq:shift_nu} and~\eqref{eq:tfm_stability}
both arise from linearizing the block map $F$ around a fixed point,
but at different points and addressing different questions.
\textbf{The first}, equation~\eqref{eq:shift_nu}, is static: given that
attention has been added, where is the new fixed point $\tilde{x}^*$?
Here $\delta$ is a fixed vector measuring the displacement from $x^*$
to $\tilde{x}^*$.
\textbf{The second}, equation~\eqref{eq:tfm_stability}, is dynamic: once at
$\tilde{x}^*$, do nearby trajectories return to it or diverge?
Here $\epsilon_n$ is an evolving perturbation whose amplitude changes
with the number of iterations $n$.
\paragraph{Invertibility of $M^*$ and the bifurcation condition:}
Formula~\eqref{eq:fp_shift} requires $M^*$ to be invertible,
i.e.\ $\nu_k \neq 0$ for all $k$.
Since the eigenvalues of $I + M^*$ are $1 + \nu_k$, a zero eigenvalue
of $M^*$ corresponds to a unit eigenvalue of $I + M^*$, precisely the
bifurcation condition: the fixed point may be destroyed, created, or
exchange stability, and the perturbative prediction breaks down since
${M^*}^{-1}$ no longer exists.
There are three cases:
\begin{itemize}
    \item If $M^*$ is invertible and $|1+\nu_k| < 1$ for all $k$:
    the new fixed point exists, is unique to first order, and is
    stable.
    \item If some $\nu_k \to 0$: $M^*$ becomes singular,
    $\|\delta\| \to \infty$, and perturbation theory breaks down.
    The system is at a bifurcation point; the fixed point may cease to
    exist or become non-unique.
    \item If $|1+\nu_k| > 1$ for some $k$: ${M^*}^{-1}$ exists but
    the new fixed point is unstable: attention has destabilized the
    fixed point in that direction.
\end{itemize}
The measurements in Experiment~1 are consistent with the second case
being realized in the long-$\xi$ regime, where the system crosses into
a qualitatively different dynamical regime that lies outside the scope
of the perturbation expansion.
\subsection{RG classification}
\label{sec:theory_rg}
The eigenvalues $\nu_k$ of $M^*$ appear in both the shift
formula~\eqref{eq:shift_nu} and the stability
equation~\eqref{eq:tfm_stability}, and together determine the RG
character of attention in direction $v_k$.
\begin{itemize}
    \item \textbf{Irrelevant} ($-2 < \nu_k < 0$, $|1+\nu_k|<1$):
    perturbations in direction $v_k$ contract at each iteration and
    are progressively integrated out, leaving no trace at the fixed
    point. The shift $[\delta]_k = -[a+b]_k/\nu_k$ is moderate ---
    of the same order as the driving term $[a+b]_k$ --- and
    $\tilde{x}^*$ is stable in this direction.
    \item \textbf{Marginal} ($\nu_k = -2$, $|1+\nu_k|=1$):
    perturbations neither grow nor decay; the shift is finite and of
    order $[a+b]_k$. The fixed point is neutrally stable in direction
    $v_k$ and perturbations persist at constant amplitude, potentially
    producing power-law rather than exponential dependence on depth.
    Whether the system drifts away depends on nonlinear terms beyond
    the first-order expansion; distinguishing marginal from irrelevant
    in practice requires measuring the scaling of
    $\Delta_\mathrm{CKA}$ with sequence length $T$, as carried out in
    Experiment~3.
    \item \textbf{Relevant} ($\nu_k > 0$ or $\nu_k < -2$,
    $|1+\nu_k|>1$): perturbations grow at each iteration and drive
    the system away from $\tilde{x}^*$ toward a new stable attractor.
    In the Wilsonian picture this mode is not integrated out but
    instead grows under coarse-graining, reorganizing the system into
    a qualitatively different representational regime.
    \item \textbf{Bifurcation} ($\nu_k = 0$, $|1+\nu_k|=1$): the
    denominator of $[\delta]_k = -[a+b]_k/\nu_k$ vanishes and the
    shift diverges --- $\tilde{x}^*$ has moved infinitely far from
    $x^*$ and cannot be located perturbatively. Simultaneously
    perturbations around $\tilde{x}^*$ neither grow nor decay. Both
    equations signal the same event: the perturbative expansion breaks
    down entirely and the system reorganizes into a qualitatively
    different attractor outside the scope of the linearized expansion.
\end{itemize}
\subsection{The network as a discrete RG phase space}
\label{sec:theory_phasespace}
In \citep{haggimani2026phase1} we established that the network
implements a discrete approximation to an RG flow: each layer
executes one coarse-graining step, and the fixed-point plateau
corresponds to the attractor of that flow.
The perturbation analysis above allows us to develop this picture
more precisely.
In a continuous RG phase space, each point represents a theory
specified by its coupling constants $(g_1, g_2, \ldots)$, and the
RG flow is a vector field describing how these couplings change when
short-distance modes are integrated out. A trajectory is the path a
theory traces through this space under repeated coarse-graining;
different initial theories give rise to different trajectories. Near
a fixed point, the linearized RG transformation has eigenvectors
that define the natural axes of the space: relevant/irrelevant
directions (positive/negative scaling dimensions) flow away
from/toward the fixed point. The direction of the trajectory is
determined by which eigendirections are excited in the initial
theory, and its rate is controlled by the magnitude of the
corresponding scaling dimension.
In our discrete setting, the role of initial conditions is played
by the input distribution. The network weights are fixed, so $F$
(the residual block) and its fixed point $\tilde{x}^*$ are
determined; the only remaining freedom is the representation
$h^{(0)}$ at the initial layer, which is set by the input. The
sequence of representations $(h^{(0)},h^{(1)},\ldots,h^{(L)})$
traces a trajectory through representation space (not phase space),
and the direction and rate at which it approaches the fixed point
are shaped by which eigendirections of $\tilde{M}^*$ are excited
in $h^{(0)}$. An eigendirection $\tilde{v}_k$ with eigenvalue
$\tilde{\nu}_k$ is excited if $h^{(0)}$ has a large component in
that direction relative to the fixed point; by the stability
equation~\eqref{eq:tfm_stability}, more strongly excited directions
take longer to contract.
The fixed-point plateau observed in \citep{haggimani2026phase1} is
the empirical signature of this convergence: the trajectory reaches
$\tilde{x}^*$ at some intermediate layer $l^*$, after which
representations change negligibly because the system is already near
the attractor. The plateau therefore marks the depth at which the
dominant eigendirections of $\tilde{M}^*$ have contracted
sufficiently, and its location depends on which eigendirections were
excited in $h^{(0)}$ and how strongly.
The analogy with RG phase space is imperfect at two levels. At the
level of different networks, the analogy is closest: each trained
network has its own weights, its own map $F$, and hence its own
fixed point $\tilde{x}^*$ and scaling dimensions (eigenvalues of
$\tilde{M}^*$). Different trained networks therefore correspond to
different theories; different points in RG phase space with
different attractors and different flows. At the level of a single
network, the weights, $F$, and $\tilde{x}^*$ are all fixed, so we
are operating at a single point in RG phase space. Different inputs
correspond not to different theories but to different initial
conditions: they place the trajectory at different starting points
$h^{(0)}$ in representation space, exciting different combinations
of eigendirections of $\tilde{M}^*$, but all flowing under the same
$F$ toward the same $\tilde{x}^*$.
It is worth noting that different inputs converging to the same
$\tilde{x}^*$ is a statement about stability, not universality.
Universality in the RG sense would require different trained
networks --- with different weights, training data, or
hyperparameters --- to share the same fixed-point structure
($\tilde{x}^*$ and the eigenvalues of $\tilde{M}^*$). That is a
much stronger claim, and one we do not make here.
Whether attention is a relevant or irrelevant operator determines
how the TFM fixed point $\tilde{x}^*$ differs from the MLP fixed
point $x^*$. When attention is relevant ($|1+\nu_k|>1$ for some
$k$), the TFM fixed point has new unstable eigendirections and the
attractor shifts to a qualitatively different fixed-point structure
relative to the MLP. When attention is irrelevant ($-2<\nu_k<0$
for all $k$), the TFM fixed point differs from $x^*$ only by a
small shift $\delta$, all eigendirections contract, and the
fixed-point structure is essentially unchanged. The bifurcation
between these two cases, driven by the near-singularity of $M^*$,
is the mechanism behind the predictions that follow.
\subsection{Predictions}
\label{sec:theory_predictions}
The perturbation analysis of Section~\ref{sec:theory_pert} and the
RG classification of Section~\ref{sec:theory_rg} together yield four
testable predictions, each grounded in the structure of
equations~\eqref{eq:fp_shift},~\eqref{eq:shift_nu}, and~\eqref{eq:tfm_stability}.
The first two concern the magnitude of the fixed-point shift $\delta$
and the stability of the new fixed point in the two regimes; the third
concerns the layer specificity of the shift; and the fourth concerns
the spectral structure of the shift across Markov eigenmodes.
For short-$\xi$ chains, tokens decorrelate within a few steps and
the optimal prediction is the stationary distribution $\pi$
everywhere. As established in \citep{haggimani2026phase1}, this is
what the MLP fixed point encodes: all token representations collapse
to a low-dimensional attractor ($\erank\approx1.8$), where
positional variation has been integrated out. The fixed-point
plateau implies $x^*$ is stable~\eqref{eq:mlp_stability}, with all
eigenvalues of $I+M$ inside the unit circle.
When attention is added, the shift $\delta = -{M^*}^{-1}(a+b)$
~\eqref{eq:fp_shift} is small if both the resolvent ${M^*}^{-1}$
is bounded and the driving term $a+b$ is small. In this
case, both conditions are plausibly satisfied. Since
representations at $x^*$ are nearly identical across positions,
$a = \Attn(x^*)$ carries no contextual signal; this leads to $D[a]$ being
small, which in turn means $M^*=(I+D[b])(I+D[a])- I\approx D[b]$. But if $a$ has a small variation, $x^*=a$ is close to $x^*$ and so $D[b]=Df(x^* +a)\approx Df(x^*)=M$, which means that $M^*$ is close to $M$ at the fixed point and inherits its eigenstructure that dictates the stability of $x^*$. So
$M^*$ remains invertible and ${M^*}^{-1}$ is bounded.  whether $a+b$ is
small cannot be guaranteed on theoretical grounds and remains an
empirical question. Once $\delta$ is small, $\tilde{M}^* \approx
I + M^*$ and the TFM fixed point inherits the stability of $x^*$.
The prediction that $\delta$ is small is therefore not a strict
consequence of the theory but a hypothesis consistent with the
structure of the short-$\xi$ fixed point, confirmed empirically in
Experiment~1.
\medskip
\begin{prediction}[Irrelevance in short-$\xi$]
\label{pred:irrelevant}
For short-$\xi$ chains, the structure of the fixed point suggests
that both the driving term $a+b$ and the resolvent ${M^*}^{-1}$
are bounded, though this is not guaranteed by the theory alone.
We predict that the TFM converges to the same loss and similar
representational geometry as the MLP, and treat this as an
empirical test of the hypothesis.
\end{prediction}
For long-$\xi$ chains the MLP leaves a residual loss gap of $+0.017$
nats because it cannot exploit cross-token structure.
At $x^*$, token representations retain positional variation from
unresolved long-range correlations, so value vectors $V = x^* W_V$
differ across positions and $a = \Attn(x^*)$ carries a genuine
contextual signal.
Furthermore, the MLP's contraction is weakest along the slow Markov
modes (large $|\lambda_k|$), since the MLP has no mechanism to
process cross-token structure in those directions.
The corresponding $\nu_k$ may therefore be close to zero, making
${M^*}^{-1}$ large in those directions. In the long-$\xi$ case,
$a = \Attn(x^*)$ carries genuine contextual signal and varies
across positions; since $f(x^*)=0$, linearizing gives $b \approx
D[b]\cdot a$, so the driving term $a + b \approx (I+D[b])a$ is of
the same order as $a$. Whether $\|a\|$ is large enough for the
shift $\delta = -{M^*}^{-1}(a+b)$~\eqref{eq:fp_shift} to be large
cannot be guaranteed on theoretical grounds; this remains an
empirical question, and the argument is therefore suggestive rather
than a strict consequence of the theory.
It is worth noting a tension in the perturbation framework.
The fixed-point shift formula~\eqref{eq:fp_shift} is derived under
the assumption that $\|\delta\|$ is small.
Under this condition $\tilde{x}^* = x^* + \delta$ is close to
$x^*$ and $D[\tilde{b}] \approx D[b]$ follows without additional
assumptions.
In the short-$\xi$ case this is empirically self-consistent:
${M^*}^{-1}$ is bounded, and the TFM converges to similar
representational geometry as the MLP (Experiment~1), consistent
with a small shift $\delta$.
In the long-$\xi$ case the assumption $\|\delta\|\ll 1$ is expected to fail: the
slow Markov modes suggest $M^*$ has near-zero eigenvalues, and
the TFM converges to a qualitatively different representational
geometry than the MLP (Experiment~1), consistent with a large
or diverging $\delta$.
The breakdown of the perturbative expansion signals that the system is driven far from the fixed-point, and the empirical observation of a phase transition in Experiment~1 is consistent with the system crossing into a different basin of attraction. Whether this constitutes a true bifurcation in the dynamical systems sense would require a more detailed analysis of the fixed-point structure beyond the linearized expansion.
\medskip
\begin{prediction}[Relevance in long-$\xi$]
\label{pred:relevant}
For long-$\xi$ chains, $a+b$ is finite and $M^*$ may have
near-zero eigenvalues, so $\delta = -{M^*}^{-1}(a+b)$ can be
large enough to push the system out of the basin of attraction of
$x^*$, driving a phase transition to a new fixed point with
qualitatively different representational geometry.
\end{prediction}
Depending on the eigenvalues $\lambda_k$ of the transition matrix
$P$, the corresponding eigenmodes $\phi_k$ encode structure at
different scales: slow modes (large $|\lambda_k|<1$, large
$\xi_k = -1/\log|\lambda_k|$) encode long-range contextual
information that persists over many steps; fast modes (small
$|\lambda_k|$, small $\xi_k$) encode short-range structure that
decorrelates quickly.
The empirical decay length $\tau_k$, fitted from
$d_k(l) \approx A_k e^{-l/\tau_k}$, measures the approximate
number of layers the network takes to suppress a perturbation
along $\phi_k$.
From equation~\eqref{eq:tfm_stability}, using the approximation
$\tilde{M}^* \approx I + M^*$, which holds near $x^*$ when $\delta$
is small, a perturbation in direction $v_k$ evolves as
$[\epsilon_l]_k = (1+\nu_k)^l[\epsilon_0]_k$, where $l$ counts
iterations of the block map $F$.
Identifying one iteration of $F$ with one network layer, an
approximation that holds when consecutive layers are approximately
homogeneous, as in the fixed-point plateau; the decay rate is
controlled by $|1+\nu_k|$: when $|1+\nu_k|$ is close to $1$, the
perturbation persists across many layers (large $\tau_k$); when
$|1+\nu_k| \ll 1$, it collapses quickly (small $\tau_k$).
The RG prediction is that $\tau_k$ should track $\xi_k$. Note
that $\phi_k \in \mathbb{R}^T$ lives in sequence space while
$v_k \in \mathbb{R}^d$ lives in representation space,
so the correspondence is not a geometric alignment between vectors
but a relationship between scalars: $\tau_k$ should increase with
$\xi_k$ across modes. Such a correspondence would
be consistent with the network having learned to match its
contraction rates to the spectral structure of the input
distribution.
On the other hand, the MLP, being position-blind, has no mechanism
to distinguish modes and should therefore mode-blind: $\tau_k$
approximately constant across $k$ regardless of $|\lambda_k|$.
\medskip
\begin{prediction}[Mode selectivity]
\label{pred:mode}
The perturbation decay length $\tau_k$, fitted from
$d_k(l) \approx A_k e^{-l/\tau_k}$ after injecting a perturbation
along eigenmode $\phi_k$ of $P$, should be monotonically increasing
in $|\lambda_k|$ for the TFM.
The MLP should be mode-blind, with $\tau_k$ approximately constant across
$k$.
\end{prediction}
The fixed-point shift formula~\eqref{eq:fp_shift} assigns a shift
$[\delta]_k = -[a+b]_k/\nu_k$ to each direction $v_k$, but does
not specify at which layer the shift is concentrated. The driving
term $a^{(l)} + b^{(l)}$ at layer $l$ depends on the token
representations $h^{(l)}$: before the network has converged to
$x^*$, token representations retain positional variation, so
$a^{(0)}$ at the first layer carries genuine contextual signal.
Once the fixed-point plateau is reached, token representations
are nearly identical across positions, $a^{(l)}$ carries little
positional signal, and the driving term contributes negligibly to
the shift.
The RG prediction is therefore that the fixed-point shift is
concentrated at the earliest layer: the TFM commits to its new
attractor in a single coarse-graining step, consistent with the
discrete nature of the RG flow.
If attention were instead distributed uniformly across layers,
the shift would be spread evenly and no single layer would
dominate; a pattern the fixed-point structure does not predict,
since the driving term $a^{(l)}+b^{(l)}$ carries genuine
positional signal only before the fixed-point plateau is reached.
\medskip
\begin{prediction}[Layer specificity]
\label{pred:layer}
The fixed-point shift $\delta$ should be concentrated at the
earliest layer $l=0$, where token representations still retain
positional variation and the driving term $a^{(0)}+b^{(0)}$
carries genuine contextual signal. Subsequent layers, operating
on the fixed-point plateau where positional variation has already
been integrated out, should contribute negligible shifts.
\end{prediction}
\section{Experiment 1: Matched MLP vs.\ TFM Comparison}
\label{sec:e1}
We train the MLP and TFM on both $\xi$ regimes with matched
hyperparameters ($V$, $d$, $L$, $T$, $B$, steps, learning rate, mask
rate, random seed, and Markov chain $P$).
The MLP has $\approx200$K parameters vs TFM's $\approx300$K owing
to its attention weights.
\subsection{Loss profiles}
\begin{table}[ht]
\centering
\begin{tabular}{lcccc}
\toprule
Run & $H(\pi)$ (nats) & Final loss & Gap & Interpretation \\
\midrule
MLP short   & 2.7696 & 2.7660 & $-0.004$ & converged to marginal \\
TFM short   & 2.7696 & 2.7674 & $-0.002$ & attention adds nothing \\
MLP long    & 2.2125 & 2.2295 & $+0.017$ & stuck at marginal \\
TFM long    & 2.2125 & 2.0180 & $-0.195$ & exploiting long-range context \\
\bottomrule
\end{tabular}
\caption{Final loss at step 10{,}000 relative to the stationary entropy
$H(\pi)$. Although the negligibly small negative gaps for MLP short and TFM short ($\leq 0.004$ nats)
are within the noise floor of the training procedure, a negative gap indicates the model has learned to exploit
contextual correlations beyond the marginal (not in RG sense) distribution $\pi$; a gap near
zero indicates convergence to the marginal.
}
\label{tab:loss}
\end{table}
\paragraph{Short-$\xi$:}
Both models converge to within $\pm0.004$ nats of $H(\pi)$; the
inter-model gap is negligible.
For large Dirichlet parameter $\alpha=10$, rows of $P$ are nearly uniform and context carries
little predictive signal, so $H(X_t\mid X_{t-1})\approx H(\pi)$.
This confirms Prediction~\ref{pred:irrelevant}.
\paragraph{Long-$\xi$:}
The MLP plateaus at $+0.017$ nats above $H(\pi)$; the TFM
reaches $-0.195$ nats below it.
Sequences are freshly sampled at each step, so the TFM's lower
loss is not overfitting: it has learned the conditional
$P(x_t\mid x_{t-1},\ldots)$ rather than the marginal $\pi$.
For $\alpha=0.05$ each row of $P$ is heavily peaked, giving
$H(X_t\mid X_{t-1})\ll H(\pi)$ as the remaining uncertainty is small; a well-trained TFM approaches
this conditional entropy.
The loss gap is consistent with Prediction~\ref{pred:relevant};
the representational evidence is presented in
Section~\ref{sec:e1_erank}.
\subsection{Rank collapse and fixed-point geometry}
\label{sec:e1_erank}
\begin{figure}[t]
  \centering
  \includegraphics[width=0.82\linewidth]{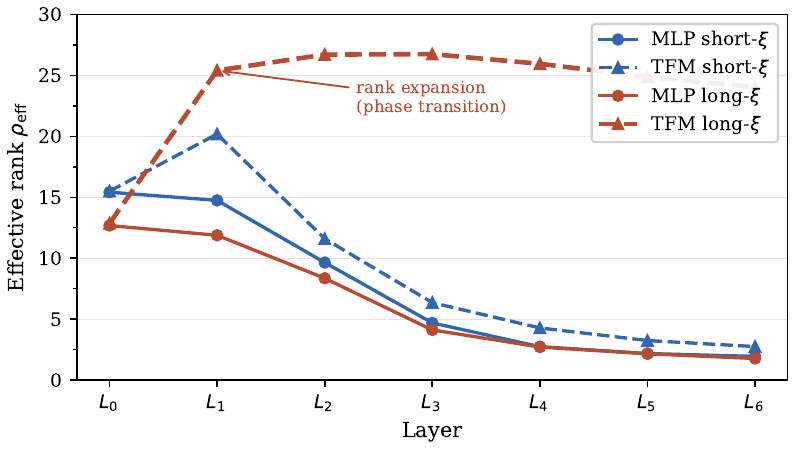}
  \caption{Effective rank $\rho_\mathrm{eff}$ across depth (flatten
    mode). \textbf{Blue}: short-$\xi$; MLP (solid) and TFM (dashed)
    both collapse monotonically. \textbf{Red solid}: MLP long-$\xi$
    compresses to $\rho_\mathrm{eff}\approx1.8$. \textbf{Red dashed}:
    TFM long-$\xi$ expands from $12.9$ to $25.4$ at $L_1$ and
    stabilizes at a high-dimensional plateau
    ($\rho_\mathrm{eff}\approx25$--$27$).}
  \label{fig:erank}
\end{figure}

Here, we examine the effective rank profiles and kernel drift
measurements to investigate how the fixed-point geometry differs
between the MLP and TFM in the two regimes, and whether the
representational signatures of Prediction~\ref{pred:relevant},
i.e.\ rank expansion and drift concentration, are observed
empirically.
\begin{table}[ht]
\centering
\begin{tabular}{lcccccc}
\toprule
Model & $L0{\to}1$ & $L1{\to}2$ & $L2{\to}3$ &
       $L3{\to}4$ & $L4{\to}5$ & $L5{\to}6$ \\
\midrule
MLP short & 0.109 & 0.355 & 0.162 & 0.085 & 0.011 & 0.002 \\
TFM short & 0.468 & 0.013 & 0.001 & 0.001 & 0.000 & 0.000 \\
\midrule
MLP long  & 0.047 & 0.247 & 0.054 & 0.003 & 0.001 & 0.000 \\
TFM long  & 0.233 & 0.060 & 0.027 & 0.028 & 0.014 & 0.024 \\
\bottomrule
\end{tabular}
\caption{Kernel drift $\Delta_l = 1 - \text{CKA}(H_l, H_{l+1})$ between
consecutive layers. The long-$\xi$ TFM does not cleanly
stabilize across depth.}
\label{tab:drift}
\end{table}
As seen in Table~\ref{tab:drift}, in the short-$\xi$ regime, MLP
drift drops to $0.011$ at $L4{\to}L5$ and $0.002$ at $L5{\to}L6$;
TFM drift reaches $0.001$ already at $L2{\to}L3$ and $0.000$
thereafter. Since both models stabilize at $L5$, we have chosen
this layer to compare the representation kernels. In the long-$\xi$
regime the TFM never cleanly stabilizes (drift $0.014$--$0.060$
through $L6$), so the $L5$ estimate is approximate for that case.
\begin{table}[ht]
\centering
\begin{tabular}{lccccccc|c|cc}
\toprule
Model & $L0$ & $L1$ & $L2$ & $L3$ & $L4$ & $L5$ & $L6$ &
ratio & CKA & rel.\ dist \\
\midrule
MLP short & 15.4 & 14.8 &  9.6 & 4.7 & 2.7 & 2.2 & 1.9 &
  $8.1\times$ & \multirow{2}{*}{0.10} & \multirow{2}{*}{0.99} \\
TFM short & 15.5 & 20.2 & 11.6 & 6.4 & 4.3 & 3.2 & 2.7 &
  $5.7\times$ & & \\
\midrule
MLP long  & 12.7 & 11.9 &  8.4 & 4.1 & 2.7 & 2.2 & 1.8 &
  $7.1\times$ & \multirow{2}{*}{0.34} & \multirow{2}{*}{17.3} \\
TFM long  & 12.9 & 25.4 & 26.7 &26.8 &26.0 &24.9 &24.0 &
  $0.5\times$ & & \\
\bottomrule
\end{tabular}
\caption{Effective rank $\erank$ across depth at the final checkpoint
(flatten mode). Compression ratio $= \erank(L_0)/\erank(L_6)$.
CKA and relative Frobenius distance compare MLP and TFM kernels
at $L5$.}
\label{tab:erank}
\end{table}
In the following discussion of rank profiles and fixed-point geometry, we refer to Table~\ref{tab:erank} and Figure~\ref{fig:erank}.
\paragraph{Short-$\xi$:}
The MLP collapses monotonically ($8.1\times$): short-range noise is
progressively integrated out with each layer.
The TFM first expands its effective rank at $L1$ ($15.5\to20.2$)
before collapsing it ($20.2\to2.7$ peak-to-end), consistent with
attention being an irrelevant perturbation: it transient
structure due to attention at $L1$ is subsequently integrated out.
Despite the geometric distance between the two final-layer kernels
(CKA $=0.10$, relative Frobenius $=0.99$), both models achieve the
same loss (Table~\ref{tab:loss}), so this geometric difference
carries no functional significance: two models can represent the
same information in geometrically unrelated ways.
\paragraph{Long-$\xi$:}
The MLP compresses the effective rank monotonically to
$\erank\approx1.8$, as in the short-$\xi$ case, confirming that
the MLP has no mechanism to exploit long-range correlations
regardless of regime.
The TFM behaves qualitatively differently: its attention mechanism
computes a distinct linear combination of value vectors for each
query position $i$,
\begin{equation}
  a_i = \sum_j \alpha_{ij} v_j, \qquad
  \alpha_{ij} = \mathrm{softmax}\!\left(\frac{q_i k_j^\top}{\sqrt{d}}\right),
\end{equation}
and since tokens carry genuinely different information across
positions, each output $a_i$ is a distinct linear combination of
value vectors rather than a near-constant vector. The outputs
$\{a_i\}$ therefore span a larger subspace than the pre-attention
representations: at layer~1, effective rank expands from $12.9$
to $25.4$.
This elevated rank stabilizes ($\erank\approx25$--$27$) across all
subsequent layers, forming a high-dimensional plateau absent in
the MLP. The representation kernels at $L5$ are geometrically
unrelated (CKA $=0.34$, relative Frobenius $=17.3$), but unlike
the short-$\xi$ case this distance is meaningful: the two models
no longer solve the same task. The TFM achieves $-0.195$ nats
below $H(\pi)$ by exploiting long-range context, while the MLP
is stuck at $+0.017$ nats above $H(\pi)$.
These are genuinely distinct attractors, not perturbatively related
fixed points. Two fixed points can be geometrically distant yet
perturbatively related if they share the same qualitative structure
(the same effective rank and the same task performance). The
long-$\xi$ fixed points share neither: the MLP collapses to
$\erank\approx1.8$ and remains near $H(\pi)$, while the TFM
stabilizes at $\erank\approx25$ and reaches $-0.195$ nats below
$H(\pi)$. These observations are consistent with the theoretical picture of
Eq.~\eqref{eq:fp_shift}: the rank expansion and loss gap indicate
a large fixed-point shift $\delta$, consistent with a loss of
perturbative control rather than a smooth perturbative shift.
Whether this is driven by a large driving term $a+b$, near-zero
eigenvalues of $M^*$, or both, cannot be determined from these
measurements alone.
\paragraph{RG interpretation:}
The two regimes map directly onto the RG classification of
Section~\ref{sec:theory_rg}: short-$\xi$ attention behaves as an
irrelevant perturbation (fixed-point shift small, MLP attractor
preserved, same loss); long-$\xi$ attention behaves as a relevant
perturbation (large shift, qualitatively different attractor,
loss gap of $0.212$ nats).
Experiments~2--4 probe the internal structure of this transition
through head ablation, scaling analysis, and perturbation decay
measurements.
\begin{figure}[t]
  \centering
  \includegraphics[width=0.82\linewidth]{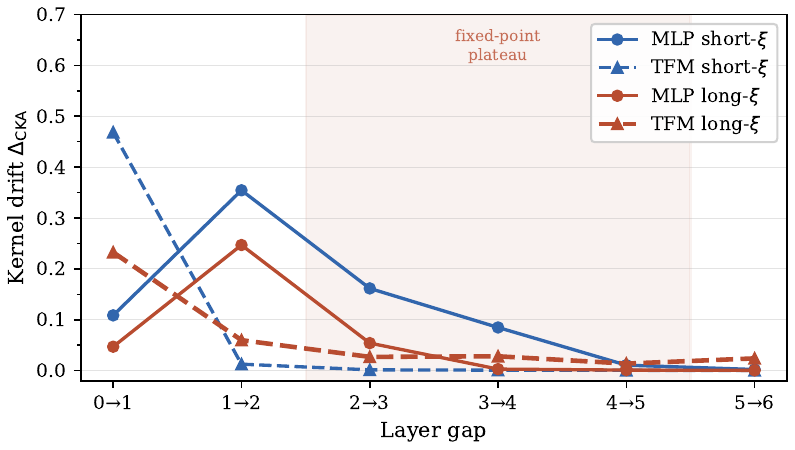}
  \caption{Kernel drift $\Delta_\mathrm{CKA}(l,l{+}1)$ (flatten
    mode). \textbf{Blue}: short-$\xi$ drift distributed across early
    layers. \textbf{Red solid}: MLP long-$\xi$ drift decays smoothly.
    \textbf{Red dashed}: TFM long-$\xi$ drift concentrates at two
    transitions, $L_0{\to}L_1$ (entry) and $L_5{\to}L_6$ (exit),
    with near-zero drift across $L_2$--$L_5$ (shaded plateau).}
  \label{fig:drift}
\end{figure}
\section{Experiment 2: Head Ablation}
\label{sec:e2}
Formula~\eqref{eq:fp_shift} describes the global fixed-point shift
when attention is added to the entire network; it cannot be
decomposed layer by layer without additional assumptions.
To understand which head (at what layer $l$) contributes most to
the shift, we introduce a heuristic local extension and write
\begin{equation}
\label{eq:layer_shift}
    \delta^{(l)} = -{M^*_l}^{-1}(a^{(l)} + b^{(l)})
    + O(\|\delta^{(l)}\|^2),
\end{equation}
where all quantities are layer specific, $a^{(l)} =
\mathrm{Attn}(h^{(l)})$, $b^{(l)} =
\mathrm{MLP}(\mathrm{LN}(h^{(l)} + a^{(l)}))$, and $M^*_l$ is
the Jacobian of $F$ at $h^{(l)}$. This is motivated
by~\eqref{eq:fp_shift} but is not implied by it.
A preliminary examination indicates which layer has the greatest
contribution to the overall shift: at $l=0$, $h^{(0)}$ reflects
raw token embeddings --- no residual block has yet processed the
input, so attention sees maximal positional variation and
$a^{(0)}+b^{(0)}$ carries the strongest contextual signal.
At $l=1$ the MLP has already begun integrating out positional
variation, so $a^{(1)}$ carries less positional signal and the
driving term contributes less to the shift.
The drift profiles (Table~\ref{tab:drift}) and rank jumps from
Experiment~1 are consistent with this picture: drift drops by a
factor of $36$ after the first layer in the short-$\xi$ TFM, and
the largest rank jump likewise occurs at $L0\to L1$ ($15.5\to20.2$
short-$\xi$, $12.9\to25.4$ long-$\xi$). These observations
motivate Prediction~\ref{pred:layer}: the attention head at layer
$0$, L0H0, should contribute the largest shift, with subsequent
heads making smaller contributions.
To test this, we ablate each attention head in turn (setting its
output to zero) and measure
\begin{equation}
  \Delta_\text{CKA}(l, h) =
  \text{CKA}(K^*_\text{MLP},\, K^*_\text{TFM,ablated}[l,h])
  - \text{CKA}(K^*_\text{MLP},\, K^*_\text{TFM,full}),
\end{equation}
at $L5$, which we chose as the reference layer where both models
have stabilized in the short-$\xi$ regime (Table~\ref{tab:drift}).
A large positive $\Delta_\text{CKA}(l,h)$ means the ablated TFM
is more similar to the MLP than the full TFM is: removing head
$(l,h)$ has brought the TFM's representations closer to the MLP
fixed point, meaning that head was responsible for driving the TFM
away from the MLP attractor.
A full RG classification (relevant, marginal, irrelevant) requires
measuring $\Delta_\text{CKA}$ at multiple values of $T$ and
fitting the scaling, which is carried out in Experiment~3; here
we report only the relative contributions.
\begin{table}[ht]
\centering
\begin{tabular}{lcc}
\toprule
Ablated head & $\Delta_\text{CKA}$ & Contribution \\
\midrule
L0H0 & $+0.119$ & \textbf{Dominant} \\
L5H0 & $+0.027$ & Moderate \\
L1H0 & $+0.025$ & Moderate \\
L3H0 & $+0.025$ & Moderate \\
L2H0 & $+0.020$ & Moderate \\
L4H0 & $+0.008$ & Negligible \\
\bottomrule
\end{tabular}
\caption{Head ablation results for the long-$\xi$ regime
($n_\text{heads}=1$, one head per layer).
Baseline CKA$(K^*_\text{MLP}, K^*_\text{TFM,full})=0.32$.
Contributions are classified by magnitude of $\Delta_\text{CKA}$
only.}
\label{tab:ablation}
\end{table}
\begin{figure}[t]
  \centering
  \includegraphics[width=0.70\linewidth]{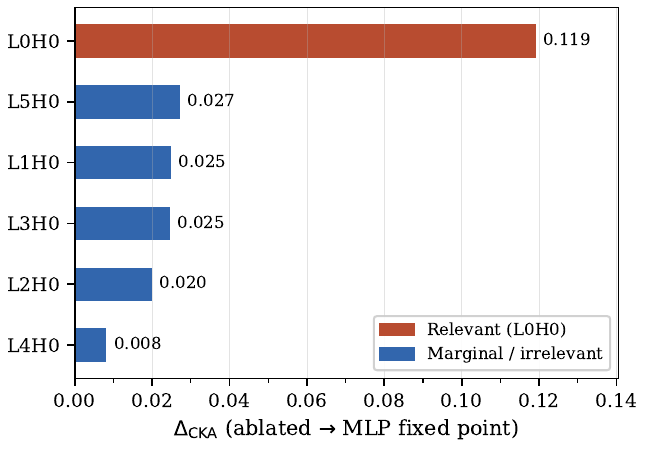}
  \caption{Head ablation $\Delta_\mathrm{CKA}(l,h)$ for the
    long-$\xi$ regime. \textbf{L0H0} (red) is the dominant
    contributor ($\Delta_\mathrm{CKA}=0.119$), more than $4\times$
    the next contribution ($0.027$, L5H0). Remaining heads (blue)
    make moderate to negligible contributions.}
  \label{fig:ablation}
\end{figure}
\paragraph{Results:}
L0H0 is strongly dominant: $\Delta_\text{CKA}=0.119$, more than
$4\times$ the next-best contribution ($0.027$,
Table~\ref{tab:ablation}). Heads L1H0, L2H0, L3H0, and L5H0
contribute $0.020$--$0.027$; L4H0 contributes $0.008$ and is
effectively negligible. This confirms Prediction~\ref{pred:layer}:
the fixed-point shift is concentrated at $l=0$, where attention
sees maximal positional variation before the MLP has begun
integrating it out. Once L0H0 has expanded effective rank at
$L0\to L1$, subsequent heads operate on the high-dimensional
plateau where remaining positional structure is already captured
and contribute negligible shifts.
This head-specialisation pattern is consistent with probing studies
that find syntactic and positional structure concentrated in early
layers \citep{voita2019analyzing, clark2019does}, though those
studies do not connect the pattern to fixed-point structure.
The broader implications for the discrete RG phase space picture
are discussed in Section~\ref{sec:disc_phasespace}.
\section{Experiment 3: Attention Scaling and RG Observables}
\label{sec:e3}
Experiments~1 and~2 characterised the fixed-point shift via
representation geometry at fixed context length $T=64$, and
identified L0H0 as the dominant contributor.
Neither experiment addressed whether any scalar statistic of the
attention weight matrices $A^{(l,h)}\in\mathbb{R}^{T\times T}$
carries an RG signature. If attention is a relevant operator in
the long-$\xi$ case, statistics of these matrices should grow in
complexity with $T$ in a way that is absent when attention is
irrelevant. The scaling exponent $\beta$, defined by
$\mathrm{stat}(T)\sim T^\beta$, captures this: a large $\beta$
signals relevance, a small $\beta$ signals irrelevance. We measure
$\beta$ for three candidate statistics across
$T\in\{16,32,64,128\}$.
\subsection{What is a genuine RG observable?}
\label{sec:observables}
We note that the statistics measured here are properties of
$A^{(l,h)}$, an intermediate quantity internal to the attention
mechanism rather than the fixed-point shift $\delta$ directly
predicted by the perturbation framework
(Section~\ref{sec:theory_pert}), so their scaling with $T$ may
reflect properties of the attention mechanism itself rather than
its effect on the residual stream. There is an inherent difficulty
here: since rows of $A^{(l,h)}$ sum to $1$ and have $T$ entries,
any statistic measuring how spread out the attention is will tend
to grow with $T$ mechanically, even for a model that has learned
to concentrate attention on a fixed number of positions.
We examine the viability of three candidate observables:
\textbf{Per-row Shannon entropy:}
For head $h$ at layer $l$, position $i$, and batch $b$, the
per-row entropy is
\begin{equation}
  H^{(l,h)}_{b,i}(T) = -\sum_{j=1}^{T} A^{(l,h)}_{b,i,j}\,
  \log A^{(l,h)}_{b,i,j},
  \qquad A^{(l,h)}_{b,i,j} =
  \frac{e^{s^{(l,h)}_{b,ij}}}{\sum_{k=1}^{T} e^{s^{(l,h)}_{b,ik}}},\quad s^{(l,h)}_{b,ij} = q^{(l,h)}_{b,i}{}^\top k^{(l,h)}_{b,j} / \sqrt{d}.
\end{equation}
It ranges from $0$ (fully peaked) to $\log T$ (uniform). We track $H(T) = \langle H^{(l,h)}_{b,i}(T)\rangle_{b,i}$ and fit
$\log H(T) = \beta_H \log T + \mathrm{const}$.
Entropy fails automatically as an RG observable because it depends only on the
attention weights, not on the value vectors at the attended
positions: two attention patterns can have identical entropy
regardless of whether the attended positions carry long-range
correlations or random noise. This is confirmed empirically in
Table~\ref{tab:e3}, where $\beta_H\approx0.27$ in both regimes.
\textbf{Participation ratio:}
For row $i$ of $A^{(l,h)}_b$, the participation ratio measures
how many tokens position $i$ effectively attends to:
\begin{equation}
  \mathrm{PR}^{(l,h,b)}_i(T)
  = \frac{1}{\sum_{j=1}^{T} \bigl(A^{(l,h)}_{b,ij}\bigr)^2}.
\end{equation}
It ranges from $1$ (fully peaked, attending to one token) to $T$
(uniform, attending equally to all tokens), with
$\mathrm{PR}^{(l,h)}(T) = \frac{1}{BT}\sum_{b,i}
\mathrm{PR}^{(l,h,b)}_i(T)$. For this to be an RG observable, one would expect $\mathrm{PR}$ to grow more
slowly with $T$ in the short-$\xi$ case (where attention has
little structure to exploit) than in the long-$\xi$ case (where
attention should spread over more informative positions as context
grows).
However, $\mathrm{PR}$ fails this test for a structural reason:
A model attending uniformly to $m$ fixed positions
out of $T$ would ideally have $\mathrm{PR}_i = m$, independent
of $T$. But softmax assigns nonzero weight to all $T$ positions;
as $T$ grows the residual weights on the remaining $T-m$ positions
do not vanish fast enough to keep $\sum_j A_{ij}^2$ constant, so
$\mathrm{PR}_i$ grows with $T$ even though the model's behaviour
has not changed. Consequently, any observed $\beta_\mathrm{PR}\approx1$, as in Table~\ref{tab:e3} for both regimes, is
a softmax artifact, not an RG signal, and the participation ratio
is disqualified as an RG observable.
\textbf{Effective rank:}
As previously defined, for $A^{(l,h)}_b \in \mathbb{R}^{T\times T}$ with singular values
$\sigma_1^{(l,h,b)} \geq \cdots \geq \sigma_T^{(l,h,b)}$, the
effective rank is
\begin{equation}
  \rho_{\mathrm{eff}}^{(l,h,b)}(T)
  = \exp\!\left(-\sum_{k=1}^{T} p_k^{(l,h,b)}
  \log p_k^{(l,h,b)}\right),
  \qquad
  p_k^{(l,h,b)} = \frac{\sigma_k^{(l,h,b)}}
  {\sum_j \sigma_j^{(l,h,b)}},
\end{equation}
with $\rho_{\mathrm{eff}}^{(l,h)}(T) = \langle
\rho_{\mathrm{eff}}^{(l,h,b)}(T)\rangle_b$.
The exponent $\beta_\rho$ measures how many effective dimensions
the attention pattern occupies as context size grows. Effective rank
avoids the problems of entropy and participation ratio: it is
computed from the singular values of $A^{(l,h)}$, which are not
constrained by softmax normalization, and captures the global
low-rank structure of the attention matrix rather than the
concentration of individual rows. For this reason, we will use effective rank as the primary RG observable, and include entropy and participation ratio only for comparison.
We fit $\log(\mathrm{stat})=\beta\log T+\mathrm{const}$ at each
layer and report results in Table~\ref{tab:e3}.
\subsection{Results}
\begin{table}[h]
\centering
\small
\begin{tabular}{llccc}
\toprule
Regime & Layer & $\beta_H$ & $\beta_{\rho}$ & $\beta_{\mathrm{PR}}$ \\
\midrule
Short-$\xi$ & L0     & 0.285 & 0.199 & 0.988 \\
            & L1     & 0.271 & 0.183 & 0.974 \\
            & L2     & 0.271 & 0.043 & 1.003 \\
            & L3--L5 & ${\sim}0.270$ & ${\sim}0.005$--$0.010$
            & ${\sim}1.000$ \\
\midrule
Long-$\xi$  & L0     & 0.269 & 0.306 & 0.892 \\
            & L1--L5 & ${\sim}0.260$--$0.266$
            & ${\sim}0.321$--$0.427$ & ${\sim}0.859$--$0.960$ \\
\bottomrule
\end{tabular}
\caption{Scaling exponents $\beta$ per layer for each statistic,
fitted from $\mathrm{stat}(T)\sim T^\beta$ across
$T\in\{16,32,64,128\}$ ($n_\mathrm{heads}=1$, all $R^2\geq0.99$).}
\label{tab:e3}
\end{table}
Entropy does not discriminate: both regimes yield
$\beta_H\approx0.27$ at every layer, confirming that entropy is
insensitive to whether attention is relevant or irrelevant.
Participation ratio shows $\beta_\mathrm{PR}\approx0.86$--$1.00$
in both regimes, consistent with the softmax artifact identified
above.
Effective rank is the only statistic that discriminates.
In the short-$\xi$ case $\beta_\rho$ falls from $0.199$ at L0 to
near zero at L3--L5; in the long-$\xi$ case it remains elevated
($0.306$--$0.427$) at all layers.
The difference $\Delta\beta_\rho\approx0.2$--$0.4$ is the clearest
RG signature in the attention patterns.
The per-layer variation in $\beta_\rho$ reflects the changing role
of attention as representations are compressed toward the fixed
point; unlike continuum RG where scaling exponents are fixed at the
fixed point, in our discrete setting they vary along the flow
because layers are not weight-tied.
Experiment~4 probes the fixed-point structure more directly by
measuring perturbation decay across Markov modes.
\section{Experiment 4: Perturbation Decay Spectrum}
\label{sec:e4}
\begin{figure}[ht]
  \centering
  \includegraphics[width=0.85\linewidth]{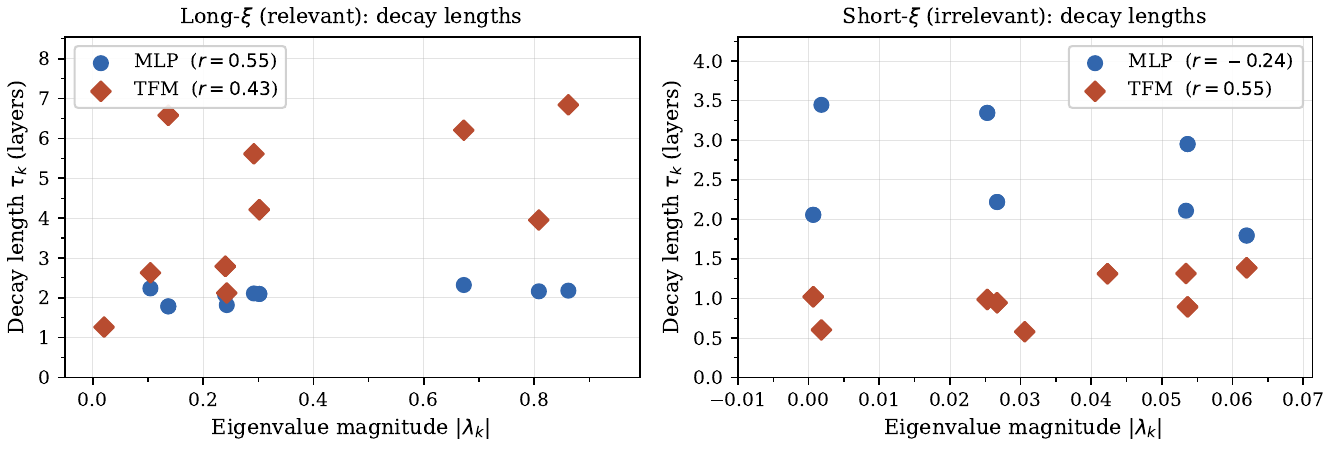}
  \caption{Perturbation decay length $\tau_k$ vs.\ eigenvalue
    magnitude $|\lambda_k|$ for each eigenmode $\phi_k$ of $P$.
    \textbf{Left (long-$\xi$):}
    MLP (blue circles, $r=0.55$) spans $\tau \in [1.78, 2.32]$
    --- narrow range, weakly selective.
    TFM (red diamonds, $r=0.44$) spans $\tau \in [1.27, 6.84]$
    --- strong selectivity, slow modes persist $5\times$ longer
    than fast modes.
    \textbf{Right (short-$\xi$):}
    ordering reverses; $\tau_\text{TFM} < \tau_\text{MLP}$ for
    every mode --- the Transformer acts as a fast integrator when
    attention is irrelevant.}
  \label{fig:spectrum}
\end{figure}

Experiment~2 showed that the network's response to attention is not
uniform across heads. Here we ask a complementary question at the
level of the input spectrum: has the network internalized the
spectral structure of the Markov chain $P$? Specifically, does it
treat slow Markov modes (which carry long-range correlations)
differently from fast modes (which carry short-range noise)?
Prediction~\ref{pred:mode} expects that it does: the Transformer
should preserve perturbations along slow modes (large $|\lambda_k|$,
long $\xi_k$) and suppress perturbations along fast modes (small
$|\lambda_k|$), because attention gives it access to the full
sequence and thus the ability to exploit long-range structure.
The MLP, which processes each token independently, should be
mode-blind, suppressing all modes at the same rate.
We test this by injecting perturbations aligned with eigenvectors
$\phi_k$ of $P$ and tracking their decay across layers.
We take a batch of sequences, represent each token as a one-hot
vector $x\in\mathbb{R}^{T\times V}$, and perturb position $t=0$, where $\epsilon=0.3$:
\begin{equation}
  x'_{0,:} = x_{0,:} + \epsilon\,\phi_k, \qquad
  x'_{t,:} = x_{t,:} \;\text{for}\; t > 0.
\end{equation}
Both $x$ and $x'$ are passed through the same trained model,
producing hidden states $h^{(l)}$ and $h'^{(l)}$ at each
layer. We track how the perturbation evolves across layers via the
cosine distance between base and perturbed representations, averaged
over the batch:
\begin{equation}
  d_k(l) = 1 - \left\langle
    \frac{h^{(l)}_{0} \cdot h'^{(l)}_{0}}
         {\|h^{(l)}_{0}\|\,\|h'^{(l)}_{0}\|}
  \right\rangle_{\!B}.
\end{equation}
We use cosine rather than Euclidean distance because residual
connections accumulate norm with depth: the base representation
$h^{(l)}_0$ grows in magnitude across layers, which would inflate
$\|h^{(l)}_0 - h'^{(l)}_0\|$ independently of whether the
perturbation is being suppressed. Cosine distance isolates the
angular separation between base and perturbed representations,
which reflects only the directional deviation of the perturbation
and is insensitive to this norm growth. A cosine distance above
$1$ would indicate that the perturbed representation has rotated
more than $90^\circ$ from the base, suggesting the perturbation
has driven the trajectory out of the linear regime; we verify that
$d_k(l)\leq1$ for all retained fits.
The stability equation~\eqref{eq:tfm_stability} predicts that
perturbations along irrelevant eigenmodes of $\tilde{M}^*$ decay
across layers. Since $\phi_k$ are eigenvectors of $P$ rather than $\tilde{M}^*$,
each perturbation along $\phi_k$ excites a mixture of network
eigenmodes (see Appendix~\ref{appendix:mode_decomp}). This is a
limitation of the experimental design: the fitted $\tau_k$
\begin{equation}
  d_k(l) \approx A_k\,e^{-l/\tau_k}
  \label{eq:decay_fit}
\end{equation}
to the observed cosine distance profile across layers
(see Appendix~\ref{appendix:decay})
is an
effective decay rate for the direction $\phi_k$ rather than a
single-eigenmode quantity, which weakens the test of the monotone
ordering predicted by Prediction~\ref{pred:mode}. A cleaner test
would inject perturbations along the eigenvectors of $\tilde{M}^*$
directly, but this is not currently tractable (see
Limitations below). We retain only
fits with $R^2>0.75$; also modes whose profiles show a flat-then-collapse
or an initial transient rise lie outside the linearized regime and
are excluded. Whether the monotone ordering of
Prediction~\ref{pred:mode} holds depends on how the learned input
projection $W$ maps Markov eigenvectors into the eigenspace of
$\tilde{M}^*$; since $W$ is determined by training rather than by
the structure of $P$ or $\tilde{M}^*$ alone, this alignment cannot
be predicted from the theory and must be verified empirically.
\begin{table}[ht]
\centering
\begin{tabular}{llcccc}
\toprule
Regime & Mode $k$ & $|\lambda_k|$ & $\xi_k$ & $\tau_\text{MLP}$
& $\tau_\text{TFM}$ \\
\midrule
Long-$\xi$  & 1    & 0.862 & 6.73 & 2.18 & 6.84 \\
            & 2    & 0.808 & 4.70 & 2.16 & 3.95 \\
            & 3    & 0.672 & 2.52 & 2.32 & 6.21 \\
            & 4--5  & 0.302 & 0.83 & 2.10 & 4.21 \\
            & 10--11 & 0.137 & 0.50 & 1.78 & 6.58 \\
            & 15    & 0.020 & 0.26 & ---  & 1.27 \\
\midrule
Short-$\xi$ & 1--2 & 0.027 & 0.28 & 2.22 & 0.95 \\
            & 3--4 & 0.062 & 0.36 & 1.79 & 1.39 \\
            & 8--9 & 0.042 & 0.32 & ---  & 1.31 \\
            & 15   & 0.002 & 0.16 & 3.45 & 0.60 \\
\bottomrule
\end{tabular}
\caption{Perturbation decay spectrum for both $\xi$ regimes
  (selected modes). $\xi_k = -1/\log|\lambda_k|$ is the
  correlation length of eigenmode $k$. Fits with $R^2 < 0.75$
  are marked~---. In the short-$\xi$ regime
  $\tau_\text{TFM} < \tau_\text{MLP}$ for every mode where both
  fits are reliable, a reversal of the long-$\xi$ ordering.}
\label{tab:spectrum}
\end{table}
\paragraph{Long-$\xi$: Transformer is spectrally selective, MLP
is not.}
In the long-$\xi$ regime, the MLP treats all modes similarly:
$\tau_\text{MLP}$ ranges only from $1.78$ to $2.32$ across all
reliable modes, a dynamic range of $1.3\times$, with Pearson
correlation $r=0.553$ between $|\lambda_k|$ and $\tau_\text{MLP}$.
The MLP suppresses slow and fast modes at nearly the same rate.
The Transformer behaves differently: $\tau_\text{TFM}$ ranges from
$1.27$ to $6.84$, a dynamic range of $5.4\times$, with $r=0.435$.
The slowest mode ($|\lambda_1|=0.862$) persists for $6.84$ layers
before decaying, while the fastest reliable mode decays within
$1.27$ layers. The Transformer strongly differentiates between
slow and fast modes, preserving the former and suppressing the
latter in a way the MLP does not (Figure~\ref{fig:spectrum}).
\paragraph{Short-$\xi$: regime reversal.}
In the short-$\xi$ regime, the eigenvalues of $P$ are small, reflecting the short-range noise carried by all modes. The central finding is that the Transformer
suppresses perturbations \emph{faster} than the MLP
($\tau_\text{TFM} < \tau_\text{MLP}$ for every mode where both fits
are reliable). This is a reversal of the long-$\xi$ ordering in which the
Transformer preserved slow modes by decaying them more slowly.
Note that despite slower modes decaying somewhat more slowly in comparison to faster ones, within the Transformer itself (Pearson correlation $r_\text{TFM} = 0.55$), this does not reflect useful selectivity: all Transformer decay times are shorter than their MLP counterparts, so the positive $r$ merely reflects that even aggressive suppression
is not perfectly uniform across modes. Mode selectivity, preserving
slow modes while discarding fast ones, requires not just
the Transformer architecture but a regime in which attention is a
relevant operator, i.e.\ one in which there is long-range structure
worth preserving. When there is none (short-$\xi$), the Transformer
uses its additional capacity to integrate out perturbations more
efficiently than the MLP rather than to be spectrally selective.
\paragraph{Limitations.}
The moderate Pearson correlations ($r_\text{MLP}=0.553$,
$r_\text{TFM}=0.435$ for long-$\xi$) indicate that the predicted monotone
ordering of $\tau_k$ with $|\lambda_k|$ is not cleanly recovered.
Two factors contribute to this issue:
First, the sample is small: with only $V-1=15$ modes and $L=6$
layers, the exponential fits carry substantial uncertainty.
Second, and more fundamentally, the injected perturbations are along $\phi_k$, the eigenvectors of $P$, not of $\tilde{M}^*$. As shown in
Appendix~\ref{appendix:mode_decomp}, perturbations along $\phi_k$
excite a \emph{mixture} of network eigenmodes simultaneously. The
fitted $\tau_k$ is an effective decay rate for this mixture, not a
property of a single eigenmode. Therefore, the $\tau_k$ values need not order monotonically
with $|\lambda_k|$ even if the underlying network eigenmodes do.
This weakens the test:
even if the network's eigenmodes are perfectly ordered by eigenvalues
$|\lambda_k|$ of $P$, the mixture means $\tau_k$ need not reflect
that ordering cleanly. Note that injecting $\phi_k$ is the
correct design for testing Prediction~\ref{pred:mode}, which
asks whether the network has internalized the spectral structure
of $P$, not an approximation to some cleaner experiment.
Nevertheless, the regime-level contrast is robust: the Transformer
is more spectrally selective than the MLP in the long-$\xi$
regime (a dynamic range of $5.4\times$ which shows that
the Transformer strongly differentiates between slow and fast modes), and more suppressive in the short-$\xi$ regime (e.g.\ modes 1--2: $\tau_\text{TFM}=0.95$
vs.\ $\tau_\text{MLP}=2.22$; mode 15: $\tau_\text{TFM}=0.60$
vs.\ $\tau_\text{MLP}=3.45$), and it
is confirmed here.
\section{Discussion}
\label{sec:discussion}
The RG framework generates four falsifiable predictions, all of which
are confirmed, albeit the fourth at the level of a regime contrast.
\textbf{Prediction~\ref{pred:irrelevant} (short-$\xi$,
irrelevance):} Experiment~1 confirms this at the level of
fixed-point geometry and task performance: both models converge
to within $\pm0.004$ nats of $H(\pi)$, CKA between final-layer
representations is $0.10$, and effective rank profiles collapse
monotonically. Experiment~4 reveals a subtler effect: attention
still contracts perturbations faster than the MLP, but without
spectral selectivity.
\textbf{Prediction~\ref{pred:relevant} (long-$\xi$, relevance):}
This is confirmed in its strongest form by CKA $=0.34$, relative Frobenius
distance $=17.3$, and rank expansion from $12.9$ to $25.4$ at
layer~1. These are not small deviations but a discontinuous
transition to a qualitatively different fixed-point structure,
consistent with the loss of perturbative control dictated by the
near-singularity of $M^*$.
Formula~\eqref{eq:fp_shift} is a first-order expansion valid when
$M^*$ stays away from singularity; in the long-$\xi$ regime this
condition fails, the resolvent diverges, and the system is driven
out of the perturbative regime entirely. Ironically, the most
dramatic confirmation of the RG prediction, \textit{a phase
transition}, is precisely the case where the perturbation formula
breaks down; a full description requires the non-perturbative RG
flow.
\textbf{Prediction~\ref{pred:layer} (layer specificity):}
L0H0 accounts for $\Delta_\text{CKA}=0.119$, more than $4\times$
the next-best contribution. The fixed-point shift is concentrated
at the first layer, where attention sees maximal positional
variation before the MLP has begun integrating it out.
This dominance is consistent with the structural prediction of
Section~\ref{sec:theory}, but admits an alternative reading: the
first-layer head sees the highest-entropy input and the most
positional variation to exploit, while later layers see partially
compressed representations. Whether L0H0 dominance is a genuine
RG effect or a structural consequence of architectural ordering
cannot be definitively distinguished here.
\textbf{Prediction~\ref{pred:mode} (mode selectivity):}
Confirmed at the regime level, with a regime reversal as the key
finding. In the long-$\xi$ regime the TFM spans
$\tau_\text{TFM}\in[1.27,6.84]$, a dynamic range of $5.4\times$
vs.\ $1.3\times$ for the MLP, selectively preserving slow Markov
modes. In the short-$\xi$ regime the TFM suppresses all modes
within one to two layers ($\tau_\text{TFM}\in[0.60,1.39]$),
faster than the MLP for every mode, with no spectral selectivity.
This reversal is a direct signature of the difference between the
relevant and irrelevant regimes: mode selectivity emerges
specifically when attention is a relevant operator.
The moderate Pearson correlations ($r_\text{MLP}=0.553$,
$r_\text{TFM}=0.435$ for long-$\xi$) are expected: as derived in
Appendix~\ref{appendix:mode_decomp}, each injected perturbation
$\phi_k$ excites a mixture of ${M}^*$ eigenmodes, so $\tau_k$
is an effective decay rate rather than a single-eigenmode quantity.
A clean monotone ordering would require $W\phi_k$ to align
predominantly with a single eigenmode of ${M}^*$ whose decay rate
tracks $|\lambda_k|$, a strong condition on learned structure that
need not hold exactly.
\subsection{RG interpretation of the results}
\label{sec:disc_phasespace}
The ablation results (Experiment~2) give a concrete picture of how
the trajectory commits to an attractor. With L0H0 active, the
trajectory is deflected at $l=0$ into the basin of the high-rank
TFM attractor; with L0H0 ablated, it falls back toward the MLP
basin. Which attractor is reached is determined at the first
coarse-graining step: the basin boundary is crossed, if at all,
in a single layer. Subsequent layers do not redirect the
trajectory (as discussed in the theory section, this is a trajectory in representation and not coupling space): the system arrives near the fixed point and each
further coarse-graining step leaves it there, consistent with the
near-zero drift observed across $L2$--$L5$ (Table~\ref{tab:drift}).
The perturbation decay results (Experiment~4) are consistent with
an RG interpretation: a network with near-marginal operators would
preserve slow Markov modes and suppress fast ones, which is what
we observe in the long-$\xi$ regime. In the short-$\xi$ regime,
all modes contract quickly regardless of their Markov eigenvalue,
consistent with all network operators being strongly irrelevant.
Together, the ablation and perturbation decay results illuminate
two complementary aspects of the RG picture. Experiment~2 shows
that which attractor the trajectory commits to is decided at the
first coarse-graining step: L0H0 either deflects the trajectory
into the high-rank basin or leaves it in the MLP basin, and
subsequent layers merely contract toward whichever fixed point
was selected. Experiment~4 shows that once near the fixed point,
the contraction structure, which operators are near-marginal
and which are strongly irrelevant, shapes how different input
distributions are processed across depth.
\section{Conclusion}
\label{sec:conclusion}
We have studied attention as a perturbation of the MLP fixed point
established in \citep{haggimani2026phase1}, using matched
architectures evaluated on synthetic Markov chain sequences with
controlled correlation length $\xi$. Four findings support the validity of
the RG interpretation.
First, attention is irrelevant for short-$\xi$ chains: both models
converge to the same loss and representational geometry, and the
fixed-point plateau structure is preserved.
Second, attention is a relevant operator for long-$\xi$ chains.
When $M^*$ has near-zero eigenvalues, the shift $\delta =
-{M^*}^{-1}(a+b)$ diverges and the system is driven out of the
perturbative regime; the observed transition to a high-dimensional
($\erank\approx25$) fixed point, with a loss $0.195$\,nats below
$H(\pi)$, is its empirical signature. The two fixed points are
geometrically unrelated (CKA $=0.34$, relative Frobenius distance
$=17.3$) and are not perturbatively connected: attention does not
produce a small correction to the MLP fixed-point structure but
drives a transition to a qualitatively different one.
Third, the dominant relevant operator is L0H0, accounting for more
than $4\times$ the representational shift of any subsequent head.
The fixed-point shift is concentrated at the first layer, where
attention sees maximal positional variation before the MLP has
begun integrating it out; subsequent heads operate on the plateau
and contribute negligible shifts.
Fourth, Experiment~4 reveals a regime reversal in perturbation
decay: in the long-$\xi$ regime the TFM selectively preserves slow
Markov modes ($\tau_\text{TFM}\in[1.27,6.84]$, a dynamic range of
$5.4\times$ vs.\ $1.3\times$ for the MLP); in the short-$\xi$
regime it suppresses all modes faster than the MLP
($\tau_\text{TFM}\in[0.60,1.39]$), with no spectral selectivity.
Together with \citep{haggimani2026phase1}, these results constitute
the first quantitative, controlled evidence that Transformers
implement qualitatively different representational dynamics
depending on the spectral structure of the input distribution, and
that a rigorous RG perturbation framework provides a predictive
account of that difference.
The present work is deliberately minimal: a single attention head,
synthetic sequences, and a controlled Markov input distribution.
In future work, we will extend the framework to full Transformer architectures
at scale, where the RG flow involves many heads, many layers, and
a rich natural-language input distribution. The phase-space picture
developed here (fixed-point attractors, relevant operators
driving transitions between fixed-point structures, irrelevant
operators integrated out without changing the attractor) provides
the conceptual scaffolding for identifying sharp depth-wise
transitions in a real language model. It remains to be seen whether these mechanisms
operate at scale.
\appendix
\renewcommand{\thesection}{Appendix~\Alph{section}}
\section{(Stability Condition for the MLP Fixed-Point)}
\label{appendix:MLP_fp}
Since each MLP block $F_\text{MLP}(x) = x+f(x)=x + \mathrm{MLP}(\mathrm{LN}(x))$, satisfies $F_\text{MLP}(x^*) = x^*$ at the fixed point, we have
\begin{equation}
    f(x^*)=0.
\end{equation}
\textbf{Stability of the MLP fixed-point:}
Linearizing the MLP block map around $x^*$ with
$x_0= x^* + \epsilon_0$, gives
\begin{equation}
    x_1 = F_\text{MLP}(x^* + \epsilon_0)
        = x^* + \epsilon_0 + f(x^* + \epsilon_0).
\end{equation}
Taylor-expanding $f$ around $x^*$ and using $f(x^*)=0$:
\begin{equation}
    f(x^* + \epsilon_0)
    = f(x^*) + D[f](x^*)\,\epsilon_0 + O(\|\epsilon_0\|^2)
    = D[f](x^*)\epsilon_0 + O(\|\epsilon_0\|^2),
\end{equation}
we get
\begin{equation}
    \epsilon_1 = (I + D[f](x^*))\,\epsilon_0,
\end{equation}
where we have defined $\epsilon_1 = x_1 - x^*$:
Iterating this map $n$ times, and defining the Jacobian $M=D[f](x^*)$:
\begin{equation}
    \epsilon_n = (I + {M})^n\,\epsilon_0,
\end{equation}
which in the eigenbasis of the Jacobian ${M}$ with eigenvalues $\mu_k$ gives
\begin{equation}
    [\epsilon_n]_k = (1 + \mu_k)^n\,[\epsilon_0]_k.
\end{equation}
The MLP fixed-point is stable if and only if all eigenvalues of
$I + {M}$ lie strictly inside the unit circle:
\begin{equation}
    |1 + \mu_k| < 1 \quad \text{for all } k.
\end{equation}
For real eigenvalues this reduces to $-2 < \mu_k < 0$: the MLP
sub-network must contract perturbations at $x^*$, but not so strongly
as to overshoot. Modes with $\mu_k > 0$ or $\mu_k < -2$ grow under iteration (unstable eigenvector directions). The fixed-point plateau observed empirically in both $\xi$ regimes implies that the trained MLP
satisfies this condition across all modes, with the low effective rank
($\erank \approx 1.8$) of the attractor reflecting that most
directions in representation space have been contracted.
\section{(Deriving the Fixed-Point Shift Formula)}
\label{appendix:fp_shift}
At the MLP fixed-point $x^*$, the MLP block $F_\text{MLP}(x) = x + f(x)$, satisfies $F_\text{MLP}(x^*) = x^*$, hence
\begin{equation}
    f(x^*)=\mathrm{MLP}(\mathrm{LN}(x^*))=0.
\end{equation}
The Transformer block
adds an attention sublayer $\Attn(x)$ before the MLP:
\begin{equation}
    F_\text{TFM}(x) = x + \Attn(x) + f\!\left(x +
    \Attn(x)\right).
\end{equation}
Suppose this addition moves the MLP fixed-point by a small amount $\delta$ to a fixed-point $\tilde{x}^*$ of the Transformer: $\tilde{x}^* = x^* + \delta$.
The transformer's fixed point condition
$F_\text{TFM}(\tilde{x}^*) = \tilde{x}^*$ implies
\begin{equation}
    \Attn(\tilde{x}^*) + f\!\left(\tilde{x}^* +
    \Attn(\tilde{x}^*)\right) = 0.
\end{equation}
Define $a \equiv \Attn(x^*)$ and expand $\Attn(\tilde{x}^*)$ to the first order in $\delta$:
\begin{equation}
    \Attn(\tilde x^*) =\Attn(x^* + \delta) = a + D[a]\delta + O(\|\delta\|^2).
\end{equation}
The argument of $f$ is therefore
\begin{equation}
    \tilde{x}^* + \Attn(\tilde{x}^*) = x^* + a +
    (I + D[a])\delta + O(\|\delta\|^2),
\end{equation}
Here, $f$ can be expanded either around $x^*$ or $x^* + a$. Our choice of the latter requires only the residual $\delta$ to be small: the attention output $a = \Attn(x^*)$ of the network is treated exactly, so the approximation remains valid regardless of the magnitude of $a$. Expanding around $x^*$ is also consistent, but requires the additional assumption that $\|a\|$ is of the same order of magnitude as $\delta$, since the error terms would include $O(\|a\|^2 + \|a\|\|\delta\| + \|\delta\|^2)$. We should also note that treating attention as a perturbation in this paper is in the dynamical systems and RG sense as we ask whether it remains relevant or irrelevant under iterations. This is a qualitative classification, not a statement that requires $\|a\|$ to be small. Our only quantitative assumption is that the new fixed-point is in the proximity of the MLP fixed-point.
Define $f(x^* + a)=b$, and expand $f$ around $x^* + a$:
\begin{equation}
    f\!\left(x^* + a + (I+D[a])\delta\right)
    = b + D[b](I+D[a])\delta + O(\|\delta\|^2).
\end{equation}
Substituting into the fixed point condition and retaining all $O(\delta)$ terms
\begin{equation}
    a + D[a]\delta + b + D[b](I+D[a])\delta = 0
\end{equation}
we get
\begin{equation}
    \bigl(a + b\bigr) +
    \bigl[D[b] + D[a]+D[b]D[a]\bigr]\delta = 0,
\end{equation}
and
\begin{equation}
\label{eq:shift}
    \delta = -{M^*}^{-1}
    \bigl(a + b\bigr).
\end{equation}
 where ${M^*}=D[b] + D[a]+D[b]D[a]$.
\section{(Stability of the TFM Fixed-Point)}
\label{appendix:tfm_stability}
The Transformer layer map
\begin{equation}
    F_\text{TFM}(x) = x + \Attn(x) + f\!\left(x +
    \Attn(x)\right)
\end{equation}
satisfies $F_\text{TFM}(\tilde{x}^*) = \tilde{x}^*$, or equivalently
$\tilde{a} + f(\tilde{x}^* + \tilde{a}) = 0$.
We apply this map at
\begin{equation}
    x_1 = F_\text{TFM}(\tilde{x}^* + \epsilon_0)=
     \tilde{x}^* + \epsilon_0
        + \Attn(\tilde{x}^* + \epsilon_0)
        + f\!\bigl(\tilde{x}^* + \epsilon_0
          + \Attn(\tilde{x}^* + \epsilon_0)\bigr),
\end{equation}
linearize the attention term around $\tilde{x}^*$, and $f$ around $\tilde{x}^* + \tilde{a}$ to get
\begin{align}
    x_1 &= \tilde{x}^* + \epsilon_0
        + \underbrace{\Attn(\tilde{x}^* + \epsilon_0)}_{
            =\,\tilde{a} + D[\tilde a]\epsilon_0 + O(\|\epsilon_0\|^2)}
        + f\!\Bigl(
            \underbrace{\tilde{x}^* + \epsilon_0
            + \Attn(\tilde{x}^* + \epsilon_0)}_{
            =\,(\tilde{x}^*+\tilde{a})
            + (I+D[\tilde a])\epsilon_0
            + O(\|\epsilon_0\|^2)}
        \Bigr) \\[6pt]
    &= \tilde{x}^* + \epsilon_0
        + \tilde{a} + D[\tilde a]\epsilon_0
        + \underbrace{f\!\bigl((\tilde{x}^*+\tilde{a})
            + (I+D[\tilde a])\epsilon_0\bigr)}_{
            =\,f(\tilde{x}^*+\tilde{a})
            +\,D[\tilde b](I+D[\tilde a])\epsilon_0
            + O(\|\epsilon_0\|^2)} \\[6pt]
    &= \tilde{x}^* + \epsilon_0
        + \tilde{a} + D[\tilde a]\epsilon_0
        + \underbrace{f(\tilde{x}^*+\tilde{a})}_{=\,-\tilde{a}}
        + D[\tilde b](I+D[\tilde a])\epsilon_0
        + O(\|\epsilon_0\|^2) \\[6pt]
    &= \tilde{x}^*
        + \bigl(I + D[\tilde a] + D[\tilde b](I+D[\tilde a])\bigr)\epsilon_0
         \\[6pt]
    &= \tilde{x}^*
        + (I+D[\tilde b])(I+D[\tilde a])\,\epsilon_0,
\end{align}
where $\tilde{a} \equiv \Attn(\tilde{x}^*)$ and $\tilde b=f(\tilde{x}^* + \tilde{a})$.
Collecting terms and defining $\epsilon_1 = x_1 - \tilde{x}^*$:
\begin{equation}
    \epsilon_1
    = \bigl(I + D[\tilde a] + D[\tilde b](I+D[\tilde a])\bigr)\epsilon_0
    = (I + D[\tilde b])(I + D[\tilde a])\,\epsilon_0,
\end{equation}
which by the same argument at each step:
\begin{equation}
    \epsilon_n = \bigl[(I+D[\tilde b])(I+D[\tilde a])\bigr]^n\,\epsilon_0=[{\tilde M}^*]^n\epsilon_0.
\end{equation}
The fixed point $\tilde{x}^*$ is stable if and only if all eigenvalues
of the composite Jacobian $(I+D[\tilde b])(I+D[\tilde a])$ lie strictly inside
the unit circle. When attention is weakly input-dependent at
$\tilde{x}^*$ (i.e.\ $\|D[\tilde a]\|\ll\|D[\tilde b]\|$), this reduces to
the condition $|1+\nu_k^*|<1$ on the eigenvalues $\nu_k^*$ of
$D[\tilde b]$, analogous to the MLP stability condition with the
Jacobian now evaluated at the attention-shifted point
$\tilde{x}^*+\tilde{a}$.
\section{(Derivation of the Exponential Decay)}
\label{appendix:decay}
From equation~\eqref{eq:tfm_stability}, a perturbation $\delta_0$ from the
fixed point $\tilde{x}^*$ evolves as
\begin{equation}
    [\delta_l]_k = (1+\nu_k)^l\,[\delta_0]_k.
\end{equation}
In the linear regime, the cosine distance $d_k(l)$ between the base
and perturbed representations is proportional to the magnitude of the
perturbation:
\begin{equation}
    d_k(l) \propto |[\delta_l]_k| = |1+\nu_k|^l\,|[\delta_0]_k|.
\end{equation}
Writing $|1+\nu_k| = e^{\log|1+\nu_k|}$:
\begin{equation}
    d_k(l) = A_k \cdot e^{l\log|1+\nu_k|},
\end{equation}
where $A_k \propto |[\delta_0]_k|$.
For an irrelevant direction $|1+\nu_k| < 1$, we have
$\log|1+\nu_k| < 0$; defining
\begin{equation}
    \tau_k = \frac{-1}{\log|1+\nu_k|} > 0,
\end{equation}
we obtain the exponential decay
\begin{equation}
\label{exp-decay}
    d_k(l) = A_k\,e^{-l/\tau_k}.
\end{equation}
Note the analogy with the Markov correlation length
$\xi_k = -1/\log|\lambda_k|$: both $\tau_k$ and $\xi_k$ are
characteristic decay scales, one for the network dynamics and one
for the input distribution.
The RG prediction that $\tau_k$ is monotonically increasing in
$|\lambda_k|$ is therefore a prediction that the network aligns its
contraction rates with the spectral structure of the input.
\section{(Mode Decomposition)}
\label{appendix:mode_decomp}
The eigenvectors $\phi_k \in \mathbb{R}^V$ of $P$ and the
eigenvectors $v_j \in \mathbb{R}^{T\times d}$ of $M^*$ live in
different spaces, so there is no reason for $\phi_k$ to align
with any single $v_j$. When we inject $\epsilon\phi_k$ at
position $0$, the perturbation in representation space is:
\begin{equation}
  \delta^{(0)} = \epsilon\,(W\phi_k,\, 0,\, \ldots,\, 0)
  \in \mathbb{R}^{T\times d}.
\end{equation}
Decomposing in the eigenbasis of $M^*$:
\begin{equation}
  \delta^{(0)} = \sum_j c_j^{(k)}\,v_j, \qquad
  c_j^{(k)} = \langle v_j,\,\delta^{(0)}\rangle,
\end{equation}
where the coefficients $c_j^{(k)}$ are generically all nonzero
since $W\phi_k$ has no reason to align with any single $v_j$.
After $l$ layers, using $\tilde{M}^*\approx I+M^*$:
\begin{equation}
  \delta^{(l)} = \sum_j c_j^{(k)}\,(1+\nu_j)^l\,v_j.
\end{equation}
This is a superposition of modes, each decaying or growing at rate
$|1+\nu_j|^l$. The $\tau_k$ fitted from~\eqref{eq:decay_fit} is
therefore an effective decay rate for the mixture, not a property
of a single eigenmode. It reduces to a single exponential
$A_k e^{-l/\tau_k}$ only if one eigenmode dominates. For instance, if
$|c_{j^*}^{(k)}|\gg|c_j^{(k)}|$ for all $j\neq j^*$
\begin{equation}
  \tau_k \approx -\frac{1}{\log|1+\nu_{j^*}|}.
\end{equation}
For a mode-selective TFM, the dominant network eigenmode $j^*$
, the index $j$ for which $|c_j^{(k)}|$ is largest when
Markov mode $k$ is injected , should satisfy
$|1+\nu_{j^*}|\approx 1$ when $|\lambda_k|$ is large (slow
Markov mode, long persistence) and $|1+\nu_{j^*}|\ll 1$ when
$|\lambda_k|$ is small (fast mode, rapid decay).
This is a  hypothesis about learned structure; there is no algebraic
guarantee, and it is exactly what Experiment~4 tests.
 

\end{document}